\documentclass[lettersize,journal]{IEEEtran}
\usepackage{amsmath,amsfonts}
\usepackage{array}
\usepackage[caption=false,font=normalsize,labelfont=sf,textfont=sf]{subfig}
\usepackage{textcomp}
\usepackage{stfloats}
\usepackage{url}
\usepackage{verbatim}
\usepackage{graphicx}
\usepackage{cite}
\usepackage{booktabs}
\usepackage{hyperref}
\usepackage{algorithmic}
\usepackage{algorithm}
\usepackage{wrapfig}
\usepackage{makecell}  
\usepackage{multirow}  
\usepackage{multicol}
\usepackage{colortbl}  
\usepackage{xcolor}  
\usepackage{pifont}
\usepackage{amssymb}

\usepackage[numbers]{natbib}

\begin{document}

\title{BiNoMaP: Learning Category-Level Bimanual Non-Prehensile Manipulation Primitives} 

\author{Huayi Zhou,~\IEEEmembership{Member,~IEEE,} Kui Jia,~\IEEEmembership{Member,~IEEE}
\thanks{This paper was produced by the IEEE Publication Technology Group. They are in Piscataway, NJ.}
\thanks{Manuscript received April 19, 2026; revised August 16, 2026. H. Zhou and K. Jia (corresponding author e-mail: kuijia@cuhk.edu.cn) are with School of Data Science, The Chinese University of Hong Kong, Shenzhen.}}

\markboth{Journal of \LaTeX\ Class Files,~Vol.~14, No.~8, August~2026}%
{Shell \MakeLowercase{\textit{et al.}}: A Sample Article Using IEEEtran.cls for IEEE Journals}

\IEEEpubid{0000--0000/00\$00.00~\copyright~2026 IEEE}

\maketitle
\begin{abstract}
Non-prehensile manipulation, encompassing ungraspable actions such as pushing, poking, pivoting, and wrapping, remains underexplored due to its contact-rich and analytically intractable nature. We revisit this problem from two perspectives. First, instead of relying on single-arm setups or favorable environmental supports (e.g., walls or edges), we advocate a generalizable dual-arm configuration and establish a suite of Bimanual Non-prehensile Manipulation Primitives (BiNoMaP). Second, departing from prevailing RL-based approaches, we propose a three-stage, RL-free framework for learning structured non-prehensile skills. We begin by extracting bimanual hand motion trajectories from egocentric video demonstrations. Since these coarse trajectories suffer from perceptual noise and morphological discrepancies, we introduce a geometry-aware post-optimization algorithm to refine them into executable manipulation primitives consistent with predefined motion patterns. To enable category-level generalization, the learned primitives are further parameterized by object-relevant geometric attributes, primarily size, allowing adaptation to unseen instances with significant shape variations. Importantly, BiNoMaP supports cross-embodiment transfer: the same primitives can be deployed on two real-world dual-arm platforms with distinct kinematic configurations, without redesigning skill structures. Extensive real-robot experiments across diverse objects and spatial configurations demonstrate the effectiveness, efficiency, and strong generalization capability of our approach.
\end{abstract}
\section{Introduction}\label{sec:intro}

\begin{figure}[t]
	\begin{center}
	\includegraphics[width=1.0\linewidth]{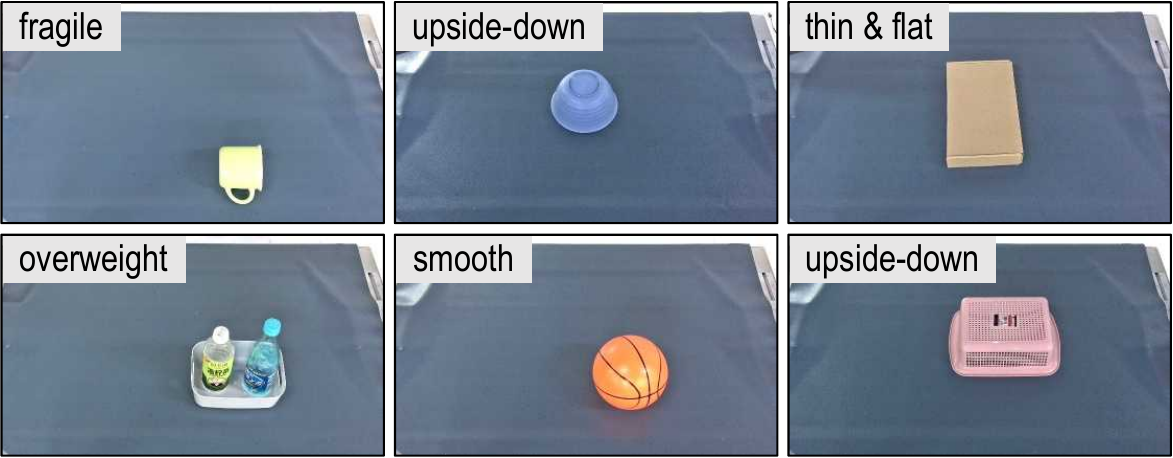}
	\vspace{-15pt}
	\caption{Examples of objects commonly encountered in daily life that cannot be directly grasped by robots. Their ungraspability may stem from their inherent geometric structure or be constrained by their current placement and orientation. To this end, we propose BiNoMaP in this paper which is precisely designed to address these non-prehensile robotic manipulation issues.} 
	\label{startCases}
	\vspace{-15pt}
	\end{center}
\end{figure}

\IEEEPARstart{N}{on-prehensil} manipulation refers to a class of robotic actions that do not rely on firm grasping but instead leverage physical interactions such as poking, or pivoting, or pushing to achieve manipulation goals \cite{zhou2019pushing, hogan2020reactive, sun2020learning, zhou2023learning, zhang2023learning}. These skills are not merely complementary to traditional grasp-based tasks; they are often essential in scenarios where grasping is physically infeasible or inefficient. In dual-arm robotic systems \cite{liu2022robot, wu2025wild, yamada2025combo, lu2025anybimanual}, non-prehensile manipulation becomes especially relevant when dealing with objects that are too fragile, too flat, or lack sufficient geometry for reliable grasping. For example, manipulating thin sheets, soft fabrics, or irregularly shaped containers often requires coordinated, contact-rich interactions such as sliding, leveraging friction, or performing indirect force applications. Some common non-prehensile examples are shown in Fig.~\ref{startCases}. These cases illustrate that grasping is not the only viable approach in robotic manipulation, and achieving robust non-prehensile manipulation demands integrating physical reasoning, multi-modal perception, and dexterous control.

Despite its importance, current non-prehensile manipulation research faces two core bottlenecks. First, most existing works operate under the simplifying assumption of a unimanual setting, often coupled with highly structured environments \cite{zhou2023hacman, cho2024corn, wu2024one, lyu2025dywa, li2025pin}. To compensate for the lack of control authority, these methods rely on artificial aids such as vertical walls, inclined planes, or boundaries to stabilize and direct object motion. However, in real-world deployments, such assumptions are rarely satisfied due to environmental unpredictability or object fragility. Consider a scenario where a thin rectangular cardboard box lies flat on a tabletop—walls and ramps are unavailable, and poking the box may damage its contents. In such cases, a more general solution is to exploit bimanual coordination \cite{krebs2022bimanual, grannen2023stabilize}, where one arm can serve as a stabilizing reference while the other executes the non-prehensile action. This configuration not only replaces inflexible external constraints with adaptive internal ones but also enables complex skills such as dual-arm wrapping \cite{grotz2024peract2, lu2025anybimanual, zhou2025you, zhou2026yoto, liu2025d} or cluttered object singulation \cite{hao2024sopedex, xu2025dexsingrasp}, which are inaccessible to single-arm systems.

\IEEEpubidadjcol
The second bottleneck lies in the heavy reliance on reinforcement learning (RL) frameworks \cite{schulman2017proximal, haarnoja2018soft, fujimoto2018addressing}. Most advanced approaches require constructing task-specific simulators that model manipulators, top-tables, and object dynamics, followed by lengthy policy training with dense environment interactions and carefully engineered reward functions. These RL pipelines are often sensitive to hyperparameter tuning and face substantial sim-to-real gaps due to inaccuracies in simulated physics, including mass distributions, contact dynamics, or friction coefficients. While recent works attempt to mitigate this gap via world models or differentiable simulators \cite{lyu2025dywa, li2025pin, huang2025particleformer}, they still inherit the inherent limitations of RL, including slow convergence and poor generalization. To the best of our knowledge, our work is the first to propose a fully RL-free paradigm for learning bimanual non-prehensile skills through imitation and geometric reasoning.

\begin{figure*}
	\begin{center}
	\includegraphics[width=1.0\linewidth]{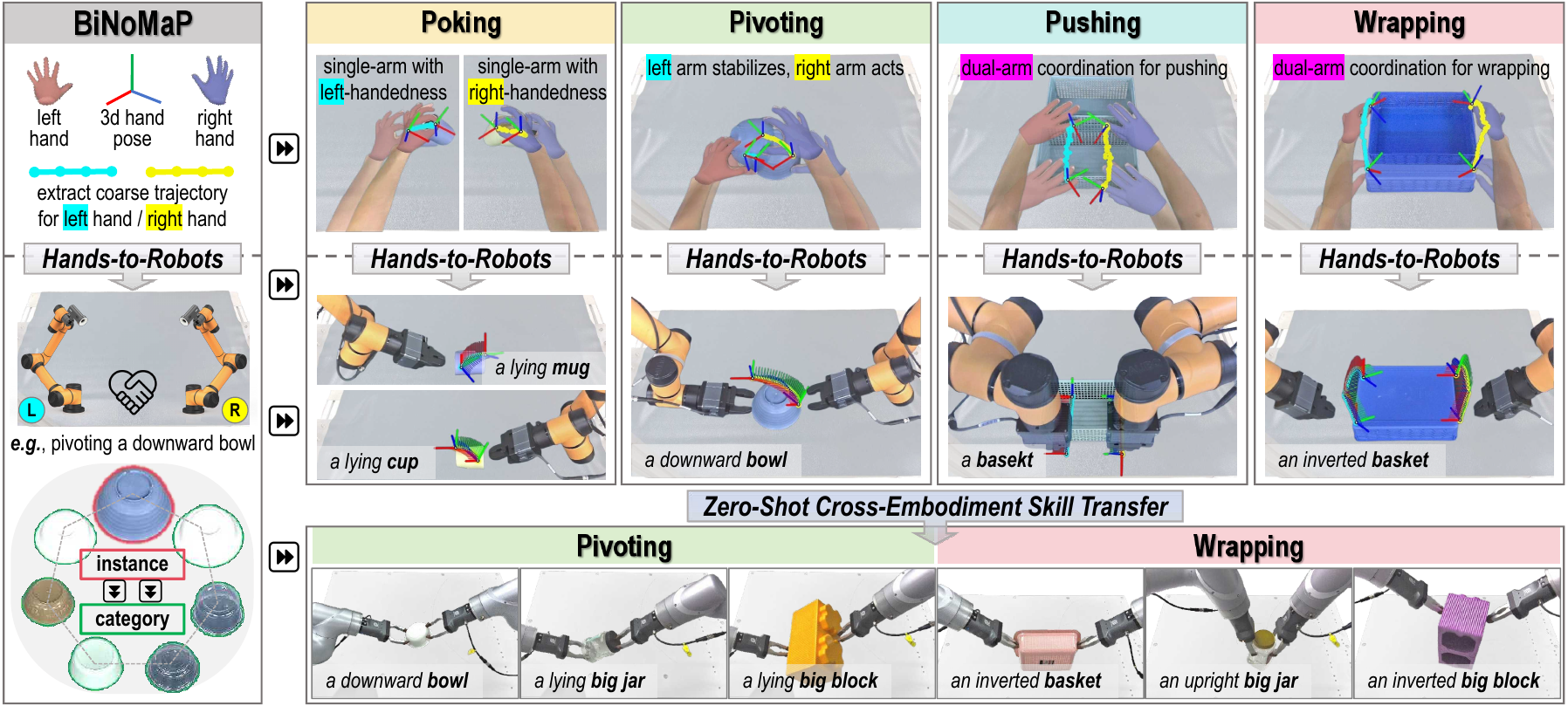}
	\vspace{-15pt}
	\caption{The illustration of proposed \textbf{Bi}manual \textbf{No}n-Prehensile \textbf{Ma}nipulation \textbf{P}rimitives (\textbf{BiNoMaP}). \textit{(Left)} We propose to extract coarse hand trajectories of bimanual non-prehensile skills from human video demonstrations, and then transfer them to the dual-arm robot. \textit{(Right)} We extensively validated BiNoMaP on four skills (e.g., poking, pivoting, pushing, and wrapping) with the embodiment-agnostic skill transfer capability.} 
	\label{teaser}
	\vspace{-15pt}
	\end{center}
\end{figure*}

To this end, we present a three-stage practical paradigm that combines hardware generality with algorithmic efficiency. We employ a dual-arm setup with parallel-jaw grippers, which not only supports unimanual non-prehensile skills but also enables more complex bimanual ones. In the \textit{\textbf{first}} stage, inspired by prior work on learning from human video demonstrations \cite{grauman2022ego4d, grauman2024ego, chen2025vividex, papagiannis2025rx}, we extract primitive bimanual motion trajectories from human hand videos for task-specific non-prehensile behaviors. Unlike grasp-based tasks, where 1$\sim$3 cm errors in hand-object alignment may be tolerable, non-prehensile tasks are often extremely sensitive to deviations: even 3$\sim$5 mm misalignment could lead to premature contact loss or over-compression, causing instability or failure. To address this, our \textit{\textbf{second}} stage introduces a geometry-aware post-optimization algorithm that leverages object shape priors to refine these noisy trajectories into smooth, task-specific motion primitives. These refined trajectories, which we term \textbf{Bi}manual \textbf{No}n-Prehensile \textbf{Ma}nipulation \textbf{P}rimitives (\textbf{BiNoMaP}), exhibit high success rates and are robust to object pose variations. The concept of primitive skills shares a similar spirit with seminal works focusing on primitive movement \cite{schaal2006dynamic, nah2024modularity}, or behaviors \cite{hogan2012dynamic, barreiros2026careful}. After that, in the \textit{\textbf{third}} stage, we further extend these primitives to unseen objects within the same category by parameterizing them with object-specific geometric attributes, such as the length and width of a box or the diameter of a sphere. This results in a family of \textit{\textbf{Parameterized Manipulation Primitives}} that are adaptive and transferable across diverse category-level object instances.

We extensively evaluate BiNoMaP on a diverse suite of dual-arm non-prehensile skills, including poking, pivoting, pushing, and wrapping (refer Fig.~\ref{teaser}). These skills are tested on objects with varying shapes, materials, and physical properties to assess robustness and geometric adaptability. To validate the effectiveness and efficiency, we compare against strong visuomotor and RL-based baselines \cite{zhao2023learning, chi2023diffusion, ze2024dp3, zhou2023hacman, cho2024corn, lyu2025dywa}, where BiNoMaP consistently achieves higher success rates across tasks. Moreover, the learned primitives can be directly deployed on two different dual-arm robotic platforms with distinct kinematic configurations, demonstrating cross-embodiment transfer without redesigning the skill structure. We further integrate BiNoMaP with high-level vision-language models (VLMs) \cite{xiao2024florence, ravi2025sam} to enable advanced behaviors such as pre-grasping under ungraspable conditions, tabletop rearrangement, and error recovery, bridging the gap between low-level contact-aware skills with high-level semantic autonomy.

To sum up, our main contributions are four-fold: (\textbf{\romannumeral 1}) We propose the first RL-free framework for learning Bimanual Non-Prehensile Manipulation Primitives directly from egocentric human video demonstrations. (\textbf{\romannumeral 2}) We introduce a parameterization scheme that enables category-level generalization of non-prehensile skills across diverse object instances. (\textbf{\romannumeral 3}) We demonstrate the effectiveness, efficiency, versatility, and generality of BiNoMaP across a variety of tasks, objects, and strong baselines. (\textbf{\romannumeral 4}) We show that the learned primitives are embodiment-agnostic and can be transferred across dual-arm platforms with different kinematic structures, validating the cross-embodiment adaptability of the proposed framework.

\section{Related Works}\label{sec:related}

\subsection{Non-Prehensile Manipulation} 
The non-prehensile manipulation has long been recognized as a crucial topic in robotic learning \cite{lynch1999dynamic, mason1999progress, zito2012two, ren2025object}, particularly for scenarios where grasping is infeasible. Prior to the rise of deep reinforcement learning (RL), traditional approaches predominantly relied on planning-based algorithms, such as graph search \cite{maeda2001planning, hou2019robust, cheng2022contact, liang2023learning} or gradient-based optimization \cite{posa2014direct, moura2022non, xu2025tracking}. However, these methods suffer from high computational cost and require precise physical priors (e.g., mass or friction coefficients), limiting their practical applicability. Recent advances have therefore shifted towards RL-based solutions, which bypass explicit planning by directly mapping sensory inputs to control actions, even in the presence of complex, contact-rich dynamics. Some works model manipulated objects using simplified geometric abstractions such as cylinders \cite{lowrey2018reinforcement}, cuboids \cite{yuan2018rearrangement, ferrandis2023nonprehensile}, or bounding boxes \cite{kim2023pre}, enabling robust hybrid force-position control but at the cost of generalization to novel shapes. Others learn control policies that can generalize across shapes, yet they typically target single primitives like pushing \cite{zhou2019pushing, zhong2026activepusher} or pivoting \cite{zhang2023learning}. More recent frameworks aim to learn a broad repertoire of non-prehensile skills under a unified RL architecture, as seen in CORN \cite{cho2024corn}, HACMan++ \cite{jiang2024hacman}, and HAMNET \cite{cho2025hierarchical}. Additionally, the use of external dexterity—such as walls, edges, or ramps \cite{yang2024learning, wu2024one, wang2025dexterous}—is a common assumption to compensate for limited control authority, while tactile sensing has been explored to infer precise contact states \cite{oller2024tactile, ferrandis2024learning, shirai2025learning}. To mitigate sim-to-real gaps, methods such as PIN-WM \cite{li2025pin} and DyWA \cite{lyu2025dywa} incorporate world models to reduce reliance on idealized observations and complete physical laws. Nonetheless, RL-based methods remain \textit{fundamentally constrained by training instability and sample inefficiency}. Moreover, all existing studies adopt \textit{a single-arm setup and heavily depend on non-generalizable environmental assumptions}. In contrast, our work challenges these limitations by introducing a more flexible and universal dual-arm setup, coupled with a three-stage RL-free learning paradigm.

\subsection{Bimanual Robotic Manipulation}
Most bimanual robotic manipulation studies often focus on graspable skills, such as cloth-folding \cite{colome2018dimensionality}, rearrangement \cite{hartmann2022long}, bagging \cite{bahety2023bag}, handover \cite{li2023efficient}, untwisting \cite{lin2024twisting}, dressing \cite{zhu2024you} and large behavior models \cite{barreiros2026careful}. For general bimanual manipulation, typical research \cite{mirrazavi2016coordinated, krebs2022bimanual, zhaodual2023afford, zhou2025vlbiman} tends to explicitly classify them into uncoordinated and coordinated according to task characteristics. Most recently, the ALOHA series \cite{zhao2023learning, fu2024mobile, zhao2024aloha} have revolutionized bimanual manipulation by dexterous teleoperating and upgrading low-cost hardwares of real-world robots. These similar works \cite{team2024octo, kim2024openvla, liu2025rdt, barreiros2026careful} implicitly train an end-to-end Vision-Language-Action (VLA) model using massive and diverse teleoperated data, expecting to get generalized robotic models. Instead of focusing on grasp-centric bimanual tasks, only a few studies address the dual-arm non-prehensile problem. For example, \cite{liu2022robot} targets a cooking scenario using the stir-fry skill. \cite{wu2025wild} employs a hybrid dual-arm combining a gripper and a suction cup to handle shelf-based object picking. \cite{yamada2025combo} introduces a master-slave coordination scheme to accomplish constrained grasping. Other works address constrained tasks such as bimanual ball lifting \cite{grotz2024peract2, lu2025anybimanual, liu2025d}. However, these approaches either focus on task-specific behaviors or rely on mixed hardware, and \textit{do not aim to study bimanual non-prehensile skills methodically}. Moreover, the non-prehensile manipulation demands fine-grained interaction control that is difficult to achieve with ordinary dual-arm teleoperation systems—particularly those lacking force or tactile feedback \cite{zhao2023learning}. As a result, collecting high-quality demonstrations for contact-intensive non-prehensile behaviors becomes impractical, which in turn hinders the scalability of data-driven imitation learning \cite{chi2023diffusion, black2025pi0}. In contrast, we propose a unified and scalable framework for learning generalizable bimanual non-prehensile skills from videos, without requiring laborious teleoperation, or deliberate hardware modifications.

\begin{figure*}
	\begin{center}
	\includegraphics[width=1.0\linewidth]{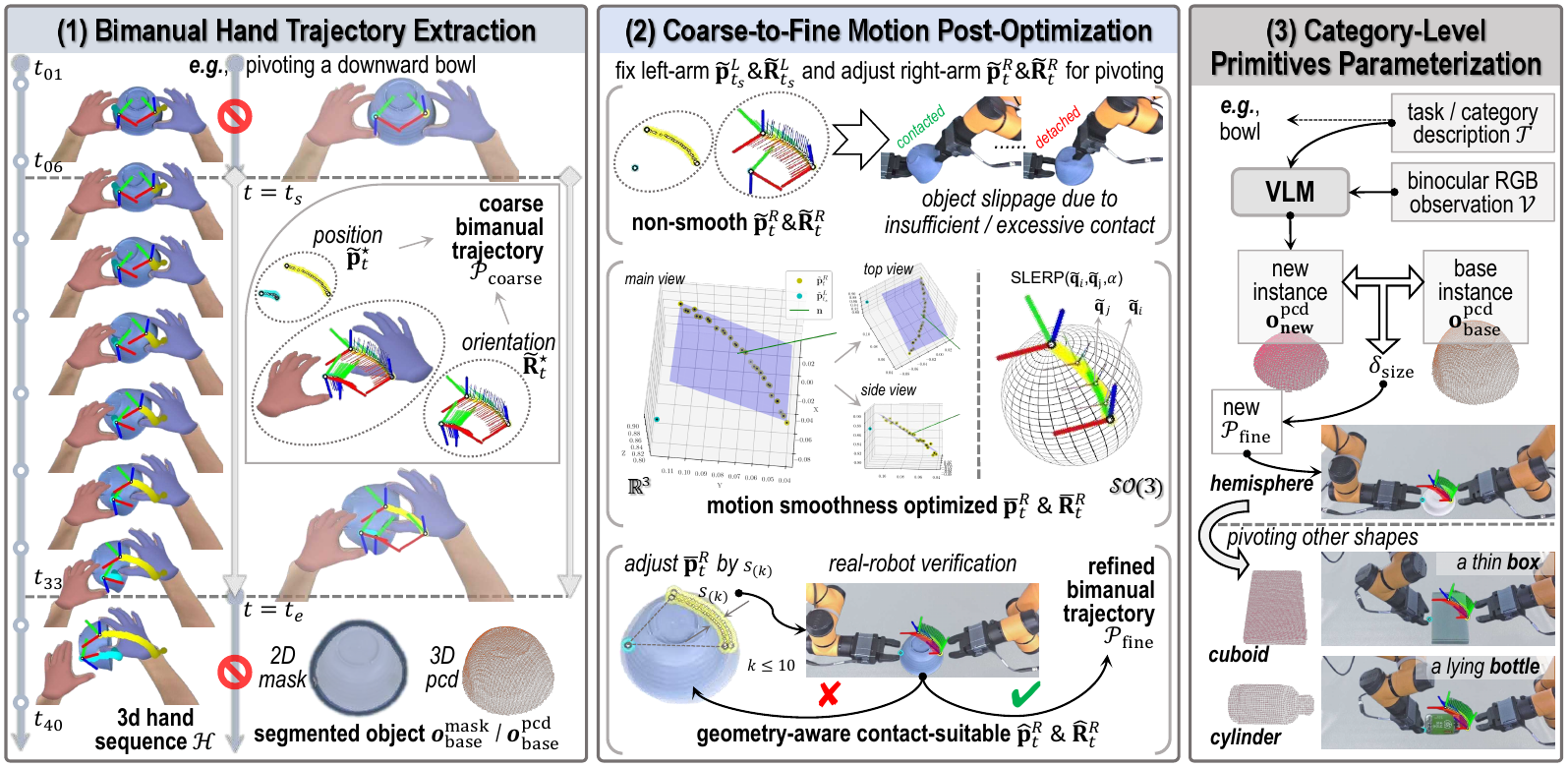}
	\vspace{-15pt}
	\caption{ The framework overview of \textbf{BiNoMaP}. (1) The first stage leverages strong priors from hand demonstrations to obtain coarse dual-arm trajectories for non-prehensile tasks. (2) The second stage refines these trajectories to mitigate multi-source noise and improve execution stability. (3) The final stage generalizes learned skills to novel objects within the same category by parameterizing primitives. }
	\label{framework}
	\vspace{-15pt}
	\end{center}
\end{figure*}

\section{Methodology}\label{sec:method}
Our proposed RL-free learning framework BiNoMaP (refer Fig.~\ref{framework}) comprises three sequential stages: \textit{Bimanual Hand Trajectory Extraction} (Sec.~\ref{BHTE}), \textit{Coarse-to-Fine Motion Post-Optimization} (Sec.~\ref{CtFMPO}), and \textit{Category-Level Primitive Parameterization} (Sec.~\ref{CLPP}). In the following, we first formalize our problem formulation and setup (Sec.~\ref{PF}), and then elaborate on three stages in detail.

\subsection{Problem Formulation}\label{PF} 

\subsubsection{Bimanual Setting and Skill Categorization} 
We target learning bimanual non-prehensile skills directly from human demonstrations without reinforcement learning, thereby bypassing limitations of RL such as slow convergence, training instability, and sim-to-real gap. Let the robot be equipped with two arms, denoted as $\mathcal{A}=\{ \mathcal{A}^L, \mathcal{A}^R \}$. We do not assume access to environmental constraints beyond manipulated objects themselves. Under this setup, we consider three major categories of non-prehensile skills: \textit{(1) single-arm skills with handedness} — tasks executable with one arm but requiring left/right preference, e.g., pushing, poking; \textit{(2) one arm manipulates, one stabilizes} — asymmetric cooperation, e.g., pivoting, occluded grasping. \textit{(3) dual-arm cooperative motion} — simultaneous symmetric/asymmetric motion of both arms, e.g., wrapping, carrying.

\subsubsection{RL-free Learning Objective} 
Learning robotic manipulation from human videos is promising yet challenging \cite{li2024okami, ko2024learning, luo2026being}, and remains largely unexplored for non-prehensile skills. Our motivation is that human bimanual hand motion inherently encodes coordination patterns required for non-prehensile manipulation. By leveraging these strong priors and focusing on error correction rather than unconstrained exploration, we can construct an efficient RL-free imitation learning paradigm. Formally, let $\mathcal{T}$ means task description (text), $\mathcal{V}$ is RGB-D video of human demonstration, $\mathcal{O}$ represents visual observation of the manipulated object(s) (e.g., RGB or 3D point cloud), and $\mathcal{A}$ indicates a dual-arm robot. We aim to learn a mapping:
\begin{equation} 
  f_{\theta} : (\mathcal{T}, \mathcal{V}, \mathcal{O}, \mathcal{A}) \longrightarrow \mathcal{P},
  \label{eqn1}
\end{equation}
where $\mathcal{P}=\{ \mathbf{p}_t, \mathbf{R}_t \}_{t=1}^T$ is the time-parameterized primitive trajectory, with $\mathbf{p}_t \in \mathbb{R}^3$ denoting the end-effector position and $\mathbf{R}_t \in SO(3)$ the end-effector orientation at time $t$. The gripper states are ignored for non-prehensile tasks. The following subsections detail how $\mathcal{P}$ is obtained.

\subsection{Bimanual Hand Trajectory Extraction}\label{BHTE}
Learning manipulation policies directly from human videos is promising yet challenging. In the \textit{vision} community, most methods focus on extracting 3D/4D hand meshes from egocentric videos \cite{grauman2022ego4d, grauman2024ego}, without addressing the cross-embodiment problem. While, in \textit{robotics}, some works use retargeting \cite{li2024okami}, keypoints \cite{wen2023any}, affordances \cite{heidinger20252handedafforder, nasiriany2025rt}, or correspondences \cite{ko2024learning} to transfer hand motion to robot end-effectors, but nearly all target prehensile tasks. Non-prehensile skills, with dense contact and low error tolerance, remain largely unexplored. Our insight is that human bimanual motion inherently encodes coordination patterns required for non-prehensile manipulation. By leveraging these strong priors and focusing on error correction rather than unconstrained exploration, we can construct an efficient imitation learning paradigm.

\subsubsection{3D Hand Reconstruction from Videos.}
We conduct human demonstrations on a table within the dual-arm workspace. For each target skill, a human demonstrator executes the motion once while a stereo camera records RGB streams, with the left view $\mathcal{V}^L$ used as the main observational source. We ensure both hands remain visible. Since only contact-relevant motion segments are necessary for primitive extraction, we manually specify start and end frames $(t_s, t_e)$ — e.g., from hands first touch the object to hands release the object — removing redundant approach/retreat phases, which can be addressed by motion planning during real robots deployment. For each frame $t \in [t_s, t_e]$ among $T'$ frames, we apply the 3D hand reconstruction algorithm WiLoR \cite{potamias2025wilor} to estimate both 3D hand shapes $\mathbf{M}^\star_t$ (the MANO \cite{romero2017embodied} parametric model capturing pose and shape parameters) and the corresponding handedness $\star \in \{ L, R \}$ (indicating left/right hand). This finally yields a hand attribute sequence $\mathcal{H} = \{ \mathbf{M}^\star_t \}_{t=t_s}^{T'}$.

\subsubsection{Hand-to-Robot Trajectory Extraction.}
To retarget hands to parallel-jaw grippers, we simplify each hand's mesh in $\mathcal{H}$ to a representative contact point: the midpoint between the thumb tip and index finger tip among 21 predefined joints by MANO. Because WiLoR outputs per-frame meshes in independent coordinates (no camera intrinsics), we thus first project the contact points to 2D image space, associate them with the 3D scene point cloud, and then transform into a unified camera coordinate frame to obtain $\mathbf{\tilde{p}}^{\star}_t \in \mathbb{R}^3$. To estimate the end-effector orientation, we approximate each hand's 3D orientation by computing a rotation matrix $\mathbf{\tilde{R}}^{\star}_t \in SO(3)$ as described in Alg.~\ref{algHandPose}, which is analogous to the eigengrasping \cite{ciocarlie2007dexterous} by aligning the index-ring fingertip direction with the gripper's spindle. The coarse bimanual trajectory is thus $\mathcal{P}_\text{coarse} = \{ (\mathbf{\tilde{p}}^{\star}_t, \mathbf{\tilde{R}}^{\star}_t) \}_{t=t_s}^{T'}$.

\begin{figure}
\vspace{-5pt}
\begin{algorithm}[H]
\caption{ Approximation of 3D Hand Pose.}  
\begin{algorithmic}\small 
	\STATE $\bullet$ \textbf{Input:} 3D hand shape $\mathbf{M}^\star_t$, 21 pre-defined 3D hand joints $I_\text{hand}$, indexs of wrist joint $i_\text{wri}$ / index-fingertip $i_\text{ind}$ / ring-fingertip $i_\text{ring}$, the given handedness $\star = L$ or $\star = R$.
	\STATE $\bullet$ \textbf{Output:} 3D hand pose $\mathbf{\tilde{R}}^{\star}_t$.  \hfill\textcolor{teal}{// either $L$ or $R$}
	\STATE Initialize $\mathbf{P}^{\star}_t \leftarrow \textbf{MANO}(\mathbf{M}^\star_t, I_\text{hand})$; \hfill\textcolor{teal}{// hand joints array}
	\STATE $p_\text{wri} \leftarrow \mathbf{P}^{\star}_t[i_\text{wri}], \; p_\text{ind} \leftarrow \mathbf{P}^{\star}_t[i_\text{ind}], \; p_\text{ring} \leftarrow \mathbf{P}^{\star}_t[i_\text{ring}]$;
	\STATE $l_\text{iw} \leftarrow (p_\text{ind} - p_\text{wri}), \; l_\text{rw} \leftarrow (p_\text{ring} - p_\text{wri})$; \hfill\textcolor{teal}{// two 3D lines}
	\STATE $v_z \leftarrow \texttt{cross\_product}(l_\text{iw}, l_\text{rw})$; \hfill\textcolor{teal}{// Z-direction}
	\STATE $\bar{v}_z \leftarrow v_z / (\texttt{normalize}(v_z) + \text{1e-8})$; \hfill\textcolor{teal}{// normalize vector }
	\STATE $v_y = l_\text{mid} \leftarrow (l_\text{iw} + l_\text{rw})/2.0$; \hfill\textcolor{teal}{// estimated Y-direction}
	\STATE $\bar{v}_y \leftarrow v_y / (\texttt{normalize}(v_y) + \text{1e-8})$; \hfill\textcolor{teal}{// normalize vector}
	\STATE $\bar{v}_x \leftarrow \texttt{cross\_product}(\bar{v}_y, \bar{v}_z)$; \hfill\textcolor{teal}{// X-direction}
	\STATE $\mathbf{\tilde{R}}^{\star}_t \leftarrow \texttt{concatenate}([\bar{v}_x, \bar{v}_y, \bar{v}_z])$; \hfill\textcolor{teal}{// $3\!\times\!3$ rotation matrix}
	\STATE \textbf{return} $\mathbf{\tilde{R}}^{\star}_t$;
\end{algorithmic}
\label{algHandPose}
\end{algorithm}
\vspace{-15pt}
\end{figure}

For later refinement, we further extract the manipulated object's geometry from the first frame $t_s$. We first utilize vision-language models (e.g., Florence-2 \cite{xiao2024florence} and SAM2 \cite{ravi2025sam}) to obtain a 2D object mask $\mathbf{o}^\text{mask}_\text{base}$. Then, we map the mask to the scene's 3D point cloud and store the segmented object point cloud $\mathbf{o}^\text{pcd}_\text{base}$ as an input for the post-optimization stage. At this point, $\mathcal{P}_\text{coarse}$ is ready for refinement to address multi-source noise from reconstruction errors, stereo matching, and hand–robot morphology gaps.

\subsection{Coarse-to-Fine Motion Post-Optimization}\label{CtFMPO}
\begin{figure*}[t]
	\begin{center}
	\includegraphics[width=1.0\linewidth]{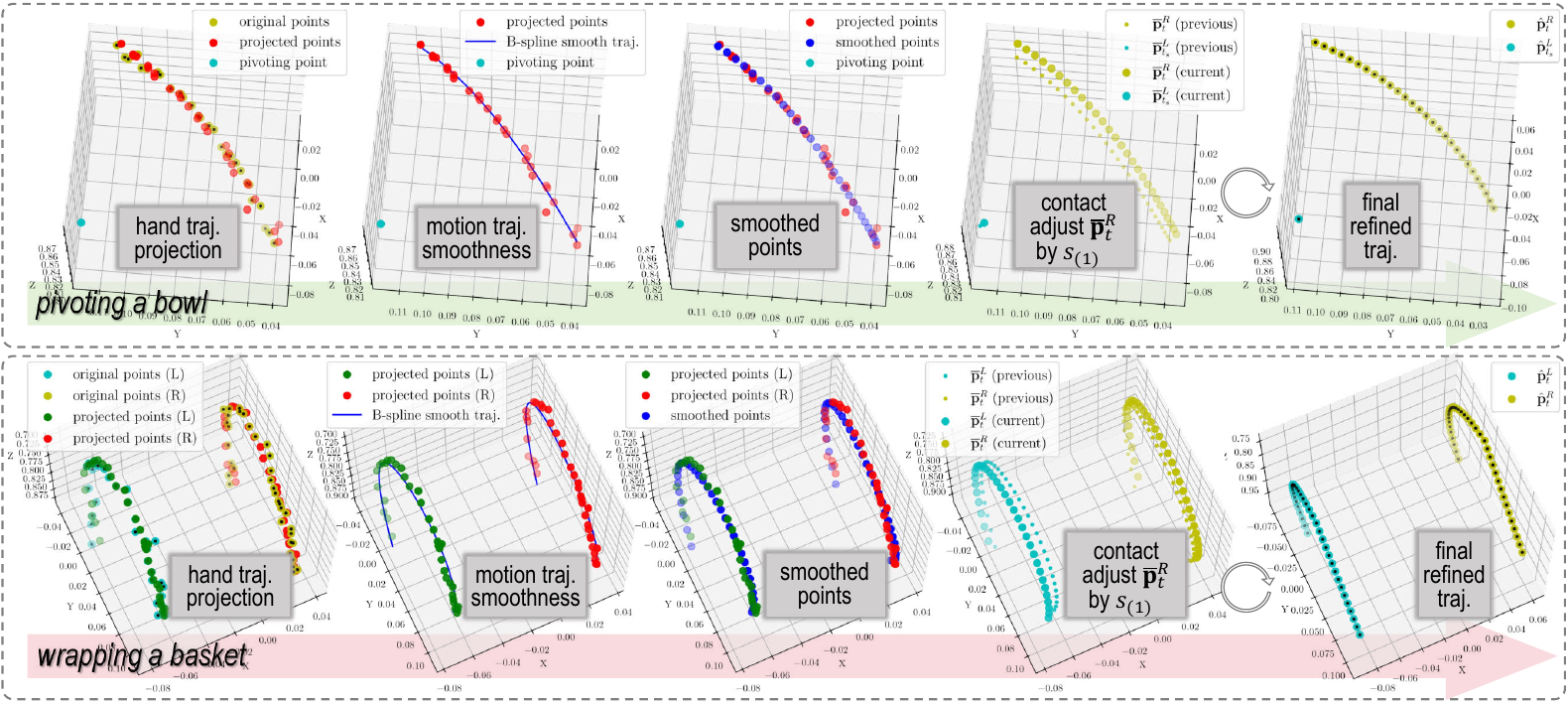}
	\vspace{-15pt}
	\caption{Illustrations of the entire trajectory point optimization process, using skills \textit{pivoting} (top) and \textit{wrapping} (down) as examples. In these two cases, we can observe how the originally irregular trajectories evolve into smoother paths with more evenly distributed waypoints. Simultaneously, the corresponding end-effector orientations become either gradually adjusted or consistently maintained. These refined spatial and rotational properties are the key factors underpinning the robustness and stability of the learned bimanual non-prehensile primitives. Best to view after zooming in.}
	\label{trajOptimize}
	\vspace{-15pt}
	\end{center}
\end{figure*}

We perform post-optimization of initially extracted $\mathcal{P}_\text{coarse}$ from two complementary perspectives: \textit{motion smoothness} and \textit{appropriate object contact}. The former reduces spatial jitter by smoothing positional and rotational transitions. The latter ensures that end-effectors maintain stable and safe contact, preventing both object slippage due to insufficient contact or rebound due to excessive force. We note an important prerequisite: all trajectory points remain \textit{coplanar} in each arm, regardless of whether two arms move synchronously or asynchronously. Overall processing illustrations of two typical skills are shown in Fig.~\ref{trajOptimize}.

\subsubsection{Motion Smoothness Optimization}
We first enforce positional smoothing with coplanarity constraint. For a single-arm trajectory $\{ \mathbf{\tilde{p}}_t \}_{t=t_s}^{T'}$, we fit an optimal plane $\prod$ to them via least squares:
\begin{equation} 
  \arg\min_{\mathbf{n},b} \sum\nolimits_{t=t_s}^{T'} (\mathbf{n}^{\top}\mathbf{\tilde{p}}_t + b)^2, \quad \texttt{s.t.} \; \| \mathbf{n} \|_2 = 1,
  \label{eqn2}
\end{equation}
where $\mathbf{n}$ is the plane normal and $b$ is the offset. All points are then orthogonally projected onto $\prod$, and trajectory smoothing filter (e.g., cubic B-spline \cite{de1978practical}) is applied in the plane coordinates to penalize curvature and enforce local smoothness.

Similarly, we smooth the sequence of rotation matrices $\{ \mathbf{\tilde{R}}_t \}_{t=t_s}^{T'}$ in $SO(3)$ for skills involving significant orientation changes. We first select a set of anchor frames $\mathcal{K} \subset \{ t_s, \cdots, t_e\}$—including the start, the end, and top-$n$ intermediate frames with minimal positional error—to serve as orientation constraints. Between anchors, we apply spherical linear interpolation (SLERP) \cite{shoemake1985animating} to ensure smooth rotational transitions aligned with the refined positions:
\begin{equation} 
\begin{aligned}
	\mathbf{\tilde{q}}(\alpha) = & \;\textsf{SLERP} (\mathbf{\tilde{q}}_i, \mathbf{\tilde{q}}_j; \alpha) = \mathbf{\tilde{q}}_i (\mathbf{\tilde{q}}_i^{\top} \mathbf{\tilde{q}}_j)^{\alpha}, \\
	& \; i, j \in \mathcal{K}, \alpha \in [0,1],
	\label{eqn3}
\end{aligned}
\end{equation}
where $\mathbf{\tilde{q}}_i$ and $\mathbf{\tilde{q}}_j$ are quaternions of matrices $\mathbf{\tilde{R}}_i$ and $\mathbf{\tilde{R}}_j$ from two selected adjacent anchors. The uniform sampling frequency of $\alpha$ depends on the number of original intermediate points omitted. The computed quaternion $\mathbf{\tilde{q}}(\alpha)$ will be converted back into a rotation matrix $\mathbf{\tilde{R}}(\alpha)$. All optimizations are performed offline, making them computationally efficient.

\subsubsection{Geometry-Aware Iterative Contact Adjustment}
After obtaining coarse yet smooth trajectories $\{ \mathbf{\overline{p}}^L_t \}_{t=t_s}^{T'}$ and $\{ \mathbf{\overline{p}}^R_t \}_{t=t_s}^{T'}$ for both arms, we refine contact geometry to ensure appropriate interaction with the object. This is achieved through an iterative, geometry-aware real-robot verification process. Without loss of generality, let the right/left arm be the primary/support arm. In iteration $k$, we first adjust the initial contact point $\mathbf{\overline{p}}^R_{t_s}$ towards the object $\mathbf{o}^\text{pcd}_\text{base}$ to achieve a target distance $d_{(k)}$. This yields an adjusted point $\mathbf{\overline{p}}^{R}_{t_s,(k)}$. This adjustment defines a scaling factor for the relative motion between two arms: $s_{(k)} = \| \mathbf{\overline{p}}^{R}_{t_s,(k)} - \mathbf{\overline{p}}^{L}_{t_s} \|_2 / \| \mathbf{\overline{p}}^{R}_{t_s} - \mathbf{\overline{p}}^{L}_{t_s} \|_2$. The scaling factor is then used to update the entire primary arm trajectory, while the support arm trajectory remains unchanged:
\begin{equation} 
  \mathbf{\overline{p}}^{R}_{t,(k)} = \mathbf{\overline{p}}^{L}_{t} + s_{(k)} ( \mathbf{\overline{p}}^{R}_{t} - \mathbf{\overline{p}}^{L}_{t} ), \quad \forall t \in [t_s, t_e].
  \label{eqn4}
\end{equation}
We initialize the verification with a large distance $d_{(1)}=$ 5mm for keeping safety. If the manipulation fails, we iteratively decrease this distance $d_{(k)} = d_{(1)} \cdot \gamma^{(k-1)} $ (e.g., $\gamma=$ 0.85) and re-evaluate, up to a maximum of ten attempts ($k \le 10$). The rationality of related hyper-parameters will be verified in ablation studies. This process, which robustly applies to diverse skills like dual-arm pivoting and wrapping, typically converges to a successful trajectory taking less than five minutes (Note that this is an one-time real-world coarse-to-fine refinement for each new object). We denote the final refined bimanual trajectory as $\mathcal{P}_\text{fine} = \{ (\mathbf{\hat{p}}^{\star}_t, \mathbf{\hat{R}}^{\star}_t) \}_{t=t_s}^{T'}$, which can be treated as an instance-level primitive skill. This approach is also significantly more efficient and stable than simulation-based RL \cite{cho2024corn, lyu2025dywa}, and safer than real-world RL \cite{luo2024serl, luo2025precise}. 

\begin{figure}[t]
	\begin{center}
	\includegraphics[width=\linewidth]{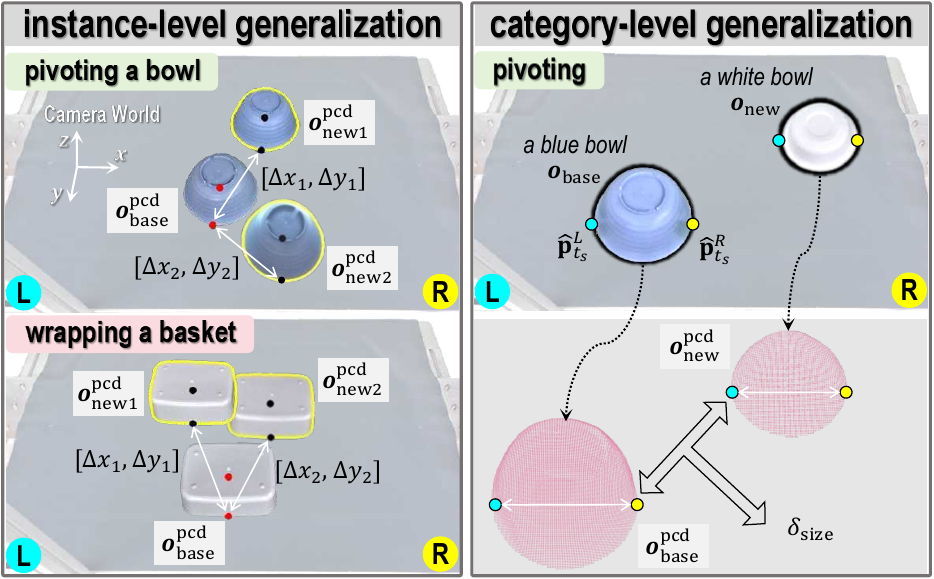}
	\vspace{-15pt}
	\caption{Details for achieving instance-level (\textit{left}) and category-level (\textit{right}) spatial generalization of learned primitive manipulation skills. These visualizations mainly serve as a detailed supplement to the right side of Fig.~\ref{framework}.}
	\label{skillAdaptation}
	\vspace{-15pt}
	\end{center}
\end{figure}

Furthermore, during deployment, we first acquire the point cloud of the object at the demonstration location, denoted as $\mathbf{o}^\text{pcd}_\text{base}$, using off the shelf VLMs \cite{xiao2024florence, ravi2025sam}. At each test time, we similarly obtain the point cloud of the same target object in its new position, denoted as $\mathbf{o}^\text{pcd}_\text{new}$. By comparing $\mathbf{o}^\text{pcd}_\text{new}$ and $\mathbf{o}^\text{pcd}_\text{base}$, we compute the in-plane displacement (e.g., $\Delta x$ and $\Delta y$), and adjust each waypoint in the primitive trajectory accordingly. This mechanism enables effective positional adaptation of the learned skill, and achieves the instance-level spatial generalization. The illustrations based on two skills \textit{pivoting} and \textit{wrapping} are shown in Fig.~\ref{skillAdaptation} left.

\subsection{Category-Level Primitive Parameterization}\label{CLPP}
To move beyond manipulating a single object instance—a common limitation of RL-based methods—we parameterize each optimized atomic trajectory to create a category-level primitive. Our geometry-aware optimization from Sec.~\ref{CtFMPO} implicitly encodes object dimensions. We make this explicit to allow the learned skill to adapt to other unseen objects with the same category. The core strategy is to treat the initially optimizated object as a base instance $\mathbf{o}^\text{pcd}_\text{base}$. For any new instance, we compute its dimensional variation relative to this base. As illustrated in Fig.~\ref{skillAdaptation} right, this is done by acquiring the new object point cloud $\mathbf{o}^\text{pcd}_\text{new}$, taking a horizontal slice at the initial contact height, and computing a characteristic dimension (e.g., the maximum point cloud distance for a bowl diameter in \textit{pivoting}):
\begin{equation} 
\begin{aligned}
 	\delta_\text{size} = &\max_{\mathbf{u}, \mathbf{v} \in \mathbf{o}^\text{pcd}_\text{new}} \| \mathbf{u} - \mathbf{v} \|_2 - \max_{\mathbf{u}, \mathbf{v} \in \mathbf{o}^\text{pcd}_\text{base}} \| \mathbf{u} - \mathbf{v} \|_2, \\
	&\quad \texttt{s.t.} \; (\mathbf{u} - \mathbf{v}) \parallel (\mathbf{\hat{p}}^L_{t_s} - \mathbf{\hat{p}}^R_{t_s}).
  \label{eqn5}
\end{aligned}
\end{equation}
The size difference $\delta_\text{size}$ between the new and base instances is then incorporated into our contact optimization Eqn.~\ref{eqn4} in a single, non-iterative step to adapt the trajectory. The corresponding verified target distance $d_{(\hat{k})}$ and scaling factor $s_{(\hat{k})}$ are modulated by $\delta_\text{size}$ to adjust the inter-arm distance, effectively resizing the manipulation primitive $\mathcal{P}_\text{fine}$ for the new object. This parameterization allows BiNoMaP to apply a learned skill to a wide range of intra-category objects without requiring new human demonstrations or repeating the full extraction and optimization pipeline. For example, in the \textit{pivoting} skill, the moving arm's trajectory is proportionally scaled, while in \textit{pushing} and \textit{wrapping}, the inter-arm distance is increased or decreased synchronously. When both category-level variation and new placements are involved, we combine this scaling procedure with the instance-level adaptation strategy to achieve robust transferability. Together, these two levels of generalization establish the scalability of BiNoMaP, enabling the learned primitives to flexibly adapt across diverse instances and object categories without retraining.

Moreover, beyond the object-level adaptation, this parameterization also enables cross-embodiment transferability across dual-arm platforms with different kinematic structures. Since the primitive $\mathcal{P}_\text{fine}$ is defined in task space through geometry-constrained relative end-effector poses rather than joint-space policies, it remains independent of robot morphology. Deployment to a new platform only requires inverse kinematics to map the adapted trajectory, without re-optimizing contact patterns or redesigning primitives. As a result, the same category-level primitive can be executed on robots with different arm lengths or base configurations, showing that BiNoMaP generalizes not only across object instances but also across embodied robotic systems. These significant advantages in scalability, convenience and versatility will be demonstrated in our experiments.

\section{Experiments}

We aim to address five central questions in our experiments. \textsf{(Q1)} Does BiNoMaP demonstrate significant advantages over state-of-the-art visuomotor-based or RL-based baselines? \textsf{(Q2)} Are all the components in the multi-phase modular design of BiNoMaP necessary and beneficial? \textsf{(Q3)} Can the proposed BiNoMaP framework generalize across a wide range of non-prehensile skills and rapidly adapt to diverse object shapes? \textsf{(Q4)} Can the learned category-level primitives in BiNoMaP generalize across dual-arm robotic platforms with different kinematics without requiring retraining or primitive redesign? \textsf{(Q5)} Can the skills learned by BiNoMaP serve as effective building blocks for accomplishing higher-level downstream long-horizon manipulation tasks?

\subsection{Experiment Setups and Protocol}

\begin{figure*}[t]
	\begin{center}
	\includegraphics[width=1.0\linewidth]{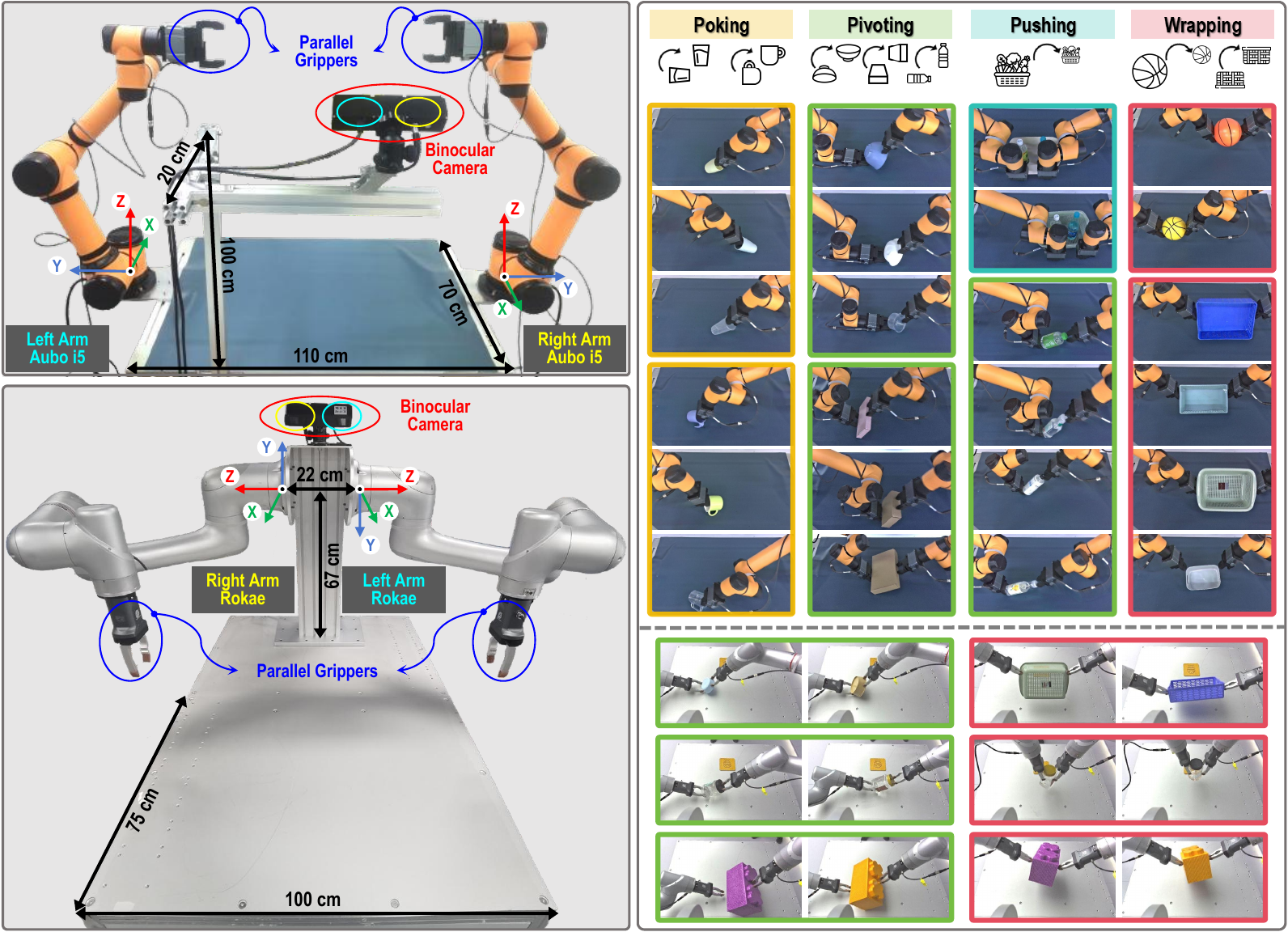}
	\vspace{-18pt}
	\caption{Illustrations of platfroms, skills and tasks. (\textit{Top-Left}) The primary fixed-base dual-arm manipulator platform used in this research. (\textit{Bottom-Left}) The another unseen humanoid dual-arm manipulator platform used for the cross-embodiment skill transfer evaluation. (\textit{Right}) Four non-prehensile skills instantiated with different manipulation tasks, diverse target objects and heterogeneous dual-arm robots.}
	\label{platforms}
	\vspace{-15pt}
	\end{center}
\end{figure*}

\subsubsection{Tasks and Setups.} We consider four typical non-prehensile skills (see illustrations in Fig.~\ref{teaser}), each of which can be instantiated on different objects to complete diverse tasks. These include \textit{uprightting a toppled cup}, \textit{flipping over an upside-down bowl}, \textit{lifting a thin cuboid box to a standing pose}, \textit{helping up a lying down bottle}, \textit{pushing a heavy basket}, \textit{translating a smooth ball}, \textit{reorienting a face-down basket to face-up}, etc. (see examples in Fig.~\ref{platforms} right). We adopt a \textbf{dual-arm setup} that eliminates the reliance on external affordances. This design leverages the intrinsic advantages of bimanual manipulation, including left–right hand complementarity, fixed-moving collaboration, and synchronous coordination. 

\subsubsection{Baselines and Metric.} To evaluate the superiority of BiNoMaP, we conduct quantitative tests on three skills (\textit{poking}, \textit{pivoting}, and \textit{wrapping}) against six baselines (including ACT \cite{zhao2023learning}, DP \cite{chi2023diffusion}, 3DP \cite{ze2024dp3}, HACMan \cite{zhou2023hacman}, CORN \cite{cho2024corn}, and DyWA \cite{lyu2025dywa}). The first three are visuomotor policies learned from real robot demonstrations, while the latter three are RL-based sim-to-real methods. We mainly compare the \textbf{Success Rates} of executing each skill on the same object placed at different positions. Unless otherwise specified, each object–task pair is evaluated with 30 real-world trials. In addition, to thoroughly evaluate the performance of BiNoMaP, we further perform various ablation studies on its modular components, as well as category-level generalization tests of each learned skill. 

\subsection{Results Comparison and Analysis}

\begin{table*}[t]\small  
	\centering
	\caption{Quantitative comparison results of our BiNoMaP and six baselines under three skills. All training times were collected on a single RTX 4090 with 24GB memory.}
	\vspace{-5pt}
	\label{tabA}
	\setlength{\tabcolsep}{7pt}
	\begin{tabular}{l|c|c|c|c|c|c|c|c|c}
	\Xhline{1.2pt}
	\multirow{2}{*}{\makecell{~\\Methods}} & \multirow{3}{*}{\makecell{Policy\\Type}} & \multirow{3}{*}{\makecell{Training\\Time (h)}} & \multicolumn{2}{c|}{\textit{poking} (L/R)} & \multicolumn{2}{c|}{\textit{pivoting} (LR)} & \multicolumn{2}{c|}{\textit{wrapping} (LR)} & \multirow{2}{*}{\makecell{Average\\Success\\Rate}} \\
	\cline{4-9}
	~ & ~ & ~ & \makecell{plastic\\cup} & \makecell{ceramic\\mug} & \makecell{plastic\\bowl} & \makecell{papery\\box} & \makecell{smooth\\ball} & \makecell{inverted\\basket} & ~ \\
	\Xhline{0.8pt} 
	ACT \cite{zhao2023learning} & Visuomotor & 8.5 & 10/30 & 03/30 & 01/30 & 05/30 & 11/30 & 01/30 & \cellcolor{gray!15}17.2\% (31/180) \\  
	DP \cite{chi2023diffusion} & Visuomotor & 4.8 & 16/30 & 11/30 & 00/30 & 10/30 & 20/30 & 04/30 & \cellcolor{gray!15}33.9\% (61/180) \\  
	DP3 \cite{ze2024dp3} & Visuomotor & 6.2 & 20/30 & 14/30 & 04/30 & 14/30 & 23/30 & 06/30 & \cellcolor{gray!15}45.0\% (81/180) \\  
	\hline
	HACMan \cite{zhou2023hacman} & RL-based & 11.1 & 07/30 & 00/30 & 01/30 & 02/30 & 07/30 & 02/30 & \cellcolor{gray!15}10.6\% (19/180) \\  
	CORN \cite{cho2024corn} & RL-based & 24.0 & 16/30 & 15/30 & 01/30 & 11/30 & 17/30 & 06/30 & \cellcolor{gray!15}36.7\% (66/180) \\  
	DyWA \cite{lyu2025dywa} & RL-based & 41.2 & 21/30 & 19/30 & 04/30 & 17/30 & 20/30 & 07/30 & \cellcolor{gray!15}48.9\% (88/180) \\  
	\hline
	BiNoMaP & RL-free & 0.1 & 29/30 & 25/30 & 20/30 & 26/30 & 30/30 & 25/30 & \cellcolor{gray!15}\textbf{86.1\% (155/180)} \\  
	\Xhline{1.2pt}
	\end{tabular}
	\vspace{-10pt}
\end{table*}

\begin{figure*}[t]
	\begin{center}
	\includegraphics[width=1.0\linewidth]{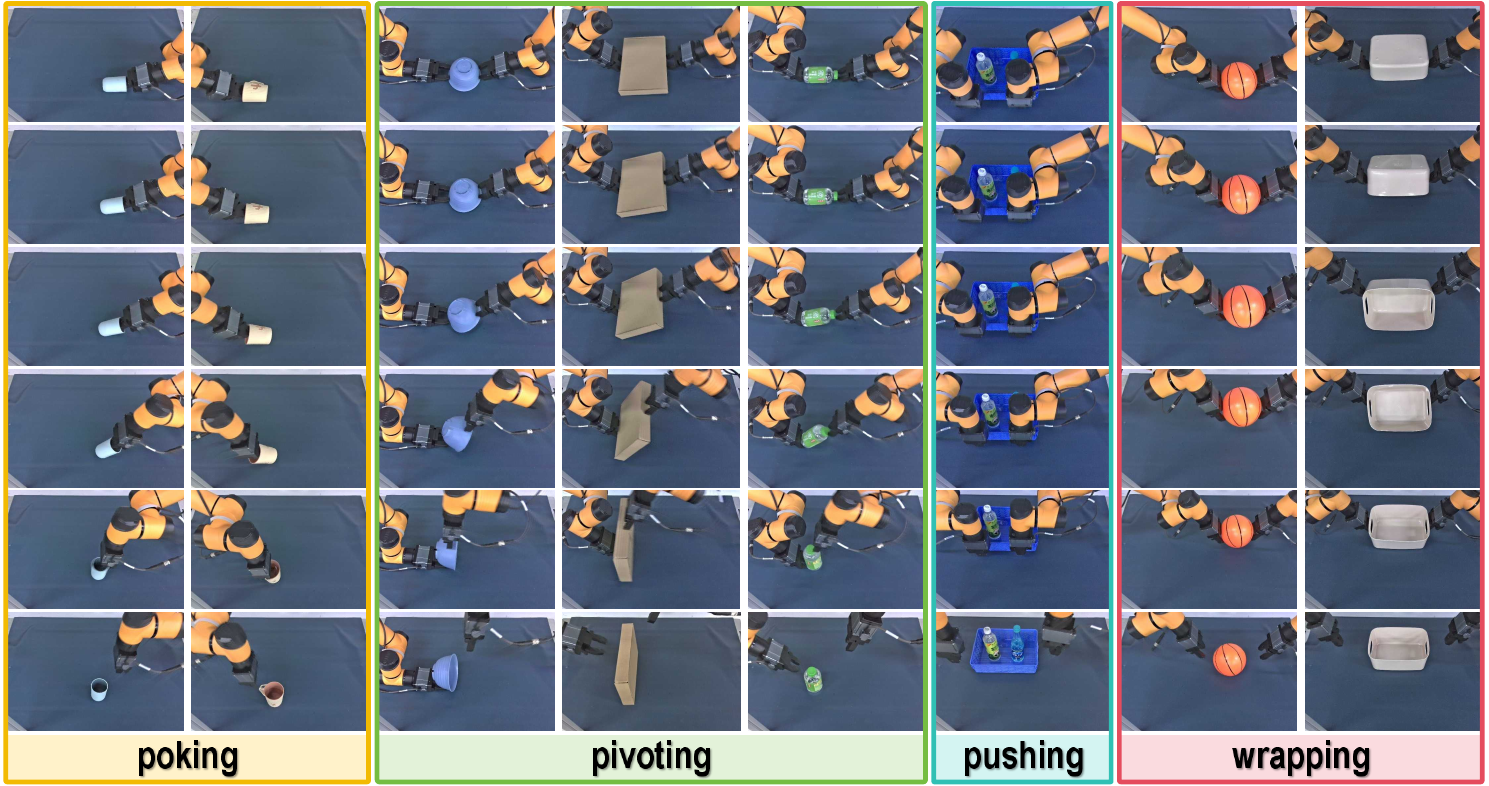}
	\vspace{-18pt}
	\caption{Qualitative real robot rollout samples of all four non-prehensile skills on the primary dual-arm manipulator platform. Best to view after zooming in.}
	\label{moreExamples}
	\vspace{-15pt}
	\end{center}
\end{figure*}

\subsubsection{Comparison with Visuomotor and RL-based Baselines \textsf{(A1)}}
As shown in Tab.~\ref{tabA}, our BiNoMaP significantly outperforms advanced non-prehensile manipulation methods across six tasks spanning three representative skills, when compared against two classes of baselines—visuomotor imitation policies and RL-based policies. Although these baselines are trained and tested on the same fixed object instances, their performance remains suboptimal. In particular, the two most challenging tasks, \textit{pivoting a plastic bowl} and \textit{wrapping an inverted basket}, require precise bimanual coordination and continuous execution of contact-rich actions. All six baselines frequently suffer from either insufficient or excessive contact, leading to premature object slippage and consequently very low success rates. For the remaining tasks, higher success rates can be partially attributed to favorable dynamics, such as leveraging friction between end-effectors and objects, or exploiting gravitational torque balance. However, these approaches either rely on direct perception-to-action mappings without explicit contact reasoning (visuomotor policies \cite{zhao2023learning, chi2023diffusion, ze2024dp3}) or incur performance degradation during sim-to-real transfer (RL-based \cite{zhou2023hacman, cho2024corn, lyu2025dywa}), and thus fail to explicitly address the requirement of frequent and fine-grained contact handling. In contrast, BiNoMaP leverages VLMs to localize novel object placements and segment 3D point clouds, followed by adaptive adjustment of learned instance-level skills. This design yields both interpretability and robustness, achieving an average success rate of \textbf{86.1\%} across all tasks—approximately twice that of latest strong baselines—demonstrating its high practicality and effectiveness (e.g., \textit{hooking the inner wall of a mug to lift it upward}, or \textit{supporting a sphere from below with both arms}).
Moreover, Fig.~\ref{moreExamples} presents more extensive qualitative results containing sequential visualizations of real-world rollouts, highlighting the continuous evolution of object states during execution. These results allow us to better appreciate the unique characteristics and challenges of non-prehensile bimanual manipulation, particularly its reliance on dense contacts and dynamic stability.

\subsubsection{Ablation Studies on the Modular Design \textsf{(A2)}}
To assess the effectiveness of BiNoMaP's modular design, we conduct ablation studies on two tasks \textit{pivoting bowl} and \textit{wrapping basket} (see Tab.~\ref{tabB}). Specifically, we evaluate the contribution of three critical components: (i) the trajectory extraction stage that approximates bimanual motion via mapping 3D hand representative points $\mapsto$ 2D pixel indices $\mapsto$ 3D object point cloud; (ii) the post-optimization stage with 3D point smoothing and 3D pose interpolation; and (iii) the iterative contact adjustment strategy. Trails are adjusted into 20. Results show that if we skip the projection-and-indexing approximation and instead rely solely on estimated 3D hand points per frame, the success rate drops drastically, even after full post-optimization—highlighting the necessity of our trajectory extraction strategy. Moreover, removing either the 3D point smoothing or 3D pose interpolation module leads to a notable performance decrease, confirming their role in eliminating jitter in both position and orientation dimensions, and thus maintaining stable contact between the object and end-effectors. Similarly, discarding the iterative contact adjustment also lowers success rates, indicating its indispensability for millimeter-level precision in contact-rich skills such as \textit{pivoting} and \textit{wrapping}, and by intuition, its utility for other non-prehensile skills as well. Furthermore, as shown in Tab.~\ref{tabC}, the hyper-parameter sweeps over the number of intermediate frames ($n$), the initial contact distance ($d_{(1)}$), and the decay factor ($\gamma$) reveal only minor fluctuations in success rates, suggesting that BiNoMaP is robust and insensitive to hyper-parameter choices, ensuring broad adaptability across different skills and objects.

\begin{table}[t]\small  
	\centering
	\caption{Ablation studies of the different four critical components in the proposed BiNoMaP.}
	\vspace{-5pt}
	\label{tabB}
	\setlength{\tabcolsep}{6pt}
	\begin{tabular}{c|c|c|c|c|c}
	\Xhline{1.2pt}
	\multicolumn{4}{c|}{Critical Components} & \multicolumn{2}{c}{Success Rate} \\
	\hline
	\makecell{map\\points} & \makecell{smooth\\points} & \makecell{smooth\\pose} & \makecell{adjust\\contact} & 
		\makecell{\textit{poking}\\bowl} & \makecell{\textit{wrapping}\\basket} \\	
	\Xhline{0.8pt} 
	\ding{55} & \ding{51} & \ding{51} & \ding{51} & 07/20 & 08/20 \\
	\ding{51} & \ding{55} & \ding{51} & \ding{51} & 10/20 & 12/20 \\
	\ding{51} & \ding{51} & \ding{55} & \ding{51} & 11/20 & 12/20 \\
	\ding{51} & \ding{51} & \ding{51} & \ding{55} & 13/20 & 14/20 \\
	\hline
	\ding{51} & \ding{51} & \ding{51} & \ding{51} & \cellcolor{gray!15}\textbf{15/20} & \cellcolor{gray!15}\textbf{17/20} \\
	\Xhline{1.2pt}
	\end{tabular}
	\vspace{-10pt}
\end{table}

\begin{table}[t]\small  
	\centering
	\caption{ Ablation studies on several hyper-parameters (taking the \textit{pivoting bowl} task as an example).}
	\vspace{-5pt}
	\label{tabC}
	\setlength{\tabcolsep}{7pt}
	\begin{tabular}{c|ccccc}
	\Xhline{1.2pt}
	top-$n$  & 1 & 2 & \textbf{3} & 4 & 5 \\
	\hline \rowcolor{gray!10}
	Success Rate & 05/10 & 06/10 & \textbf{07/10} & \textbf{07/10} & 06/10 \\
	\Xhline{0.8pt}
	$d_{(1)}$(mm)  & 3 & 4 & \textbf{5} & 6 & 7 \\ 
	\hline \rowcolor{gray!10}
	Success Rate & 06/10 & 06/10 & \textbf{07/10} & 05/10 & 05/10 \\
	\Xhline{0.8pt}
	factor $\gamma$  & 0.75 & 0.80 & \textbf{0.85} & 0.90 & 0.95 \\ 
	\hline \rowcolor{gray!10}
	Success Rate & 06/10 & \textbf{07/10} & \textbf{07/10} & 06/10 & 05/10 \\
	\Xhline{1.2pt}
	\end{tabular}
	\vspace{-10pt}
\end{table}

\begin{table*}[t]\small  
	\centering
	\caption{Quantitative results of BiNoMaP and baselines on all four skills. Trails are adjusted into 10 for each category-level instance under those \textit{unseen} objects testing.}
	\vspace{-5pt}
	\label{tabD}
	\setlength{\tabcolsep}{5pt}
	\begin{tabular}{l|c|c|c|c|c|c|c|c|c|c}
	\Xhline{1.2pt}
	Methods & \makecell{Generalization} & \multicolumn{2}{c|}{\makecell{\textit{poking}\\ (L/R)}} & \multicolumn{3}{c|}{\makecell{\textit{pivoting}\\ (LR)}} & \multicolumn{1}{c|}{\makecell{\textit{pushing}\\ (LR)}} & \multicolumn{2}{c|}{\makecell{\textit{wrapping}\\ (LR)}} & \multirow{2}{*}{\makecell{Average\\Success\\Rate}} \\
	\cline{1-10}
	~ & \makecell{Instance-Level} & \makecell{plastic\\cup} & \makecell{ceramic\\mug} & \makecell{plastic\\bowl} & \makecell{papery\\box}  & \makecell{plastic\\bottle} & \makecell{heavy\\basket} & \makecell{smooth\\ball} & \makecell{inverted\\basket} & ~ \\
	\Xhline{0.8pt} \rowcolor{gray!10}
	BiNoMaP & \textit{seen} & 29/30 & 25/30 & 20/30 & 26/30 & 25/30 & 28/30 & 30/30 & 25/30 & \textbf{86.7\%} \\  
	\Xhline{1.2pt} 
	~ & \makecell{Category-Level} & \makecell{other\\6 cups} & \makecell{other\\6 mugs} & \makecell{other\\7 bowls} & \makecell{other\\7 boxes} & \makecell{other 7\\bottles}  & \makecell{other 5\\baskets} & \makecell{other\\2 balls} & \makecell{other 5\\baskets} & ~ \\
	\Xhline{0.8pt}
	DP \cite{chi2023diffusion} & \textit{unseen} & 18/60 & 11/60 & 00/70 & 14/70 & --- & --- & 07/20 & 03/50 & 16.1\%$_{(17.8\%\downarrow)}$ \\  
	DP3 \cite{ze2024dp3} & \textit{unseen} & 26/60 & 17/60 & 03/70 & 22/70 & --- & --- & 11/20 & 05/50 & 25.5\%$_{(19.5\%\downarrow)}$ \\  
	CORN \cite{cho2024corn} & \textit{unseen} & 21/60 & 14/60 & 03/70 & 15/70 & --- & --- & 07/20 & 04/50 & 19.4\%$_{(17.3\%\downarrow)}$ \\  
	DyWA \cite{lyu2025dywa} & \textit{unseen} & 30/60 & 20/60 & 06/70 & 26/70 & --- & --- & 10/20 & 06/50 & 29.7\%$_{(18.6\%\downarrow)}$ \\  
	\hline \rowcolor{gray!10}
	BiNoMaP & \textit{unseen}  & 51/60 & 43/60 & 46/70 & 55/70 & 51/70 & 43/50 & 17/20 & 37/50 & \textbf{76.2\%$_{(10.1\%\downarrow)}$} \\  
	\Xhline{1.2pt}
	\end{tabular}
	\vspace{-10pt}
\end{table*}

\subsubsection{Generalization Across Skills and Objects \textsf{(A3)}}
To evaluate the generalizability, we conduct real-world experiments across four non-prehensile skills and a total of eight object-task pairs (see Tab.~\ref{tabD}, Fig.~\ref{platforms} and Fig.~\ref{moreExamples}). These results show that BiNoMaP applies broadly to diverse non-prehensile manipulation skills: under \textit{instance-level generalization} (same object, varying placements), it can achieve an average success rate of nearly \textbf{86.7\%}, covering objects with cylindrical, hemispherical, cuboidal, and spherical geometries, as well as materials with different frictional properties such as plastic, ceramic, and paperboard. More importantly, under the more challenging \textit{category-level generalization} setting (varying placements and object sizes), BiNoMaP still maintains a promising success rate of around \textbf{76.2\%}, with only a modest performance drop. In contrast, existing visuomotor and RL-based baselines perform poorly even in instance-level generalization (see 30$\sim$50\% success rate in Tab.~\ref{tabA}), and inevitably deteriorate further under out-of-distribution category-level tests. While such limitations could be partially alleviated by collecting more demonstrations or prolonging training in simulation, these approaches are costly, difficult to scale, and not necessarily effective for contact-rich non-prehensile tasks. By comparison, BiNoMaP offers a more \textit{scalable and deployment-friendly} solution for real-world applications.

\begin{figure}[t]
	\begin{center}
	\includegraphics[width=1.0\linewidth]{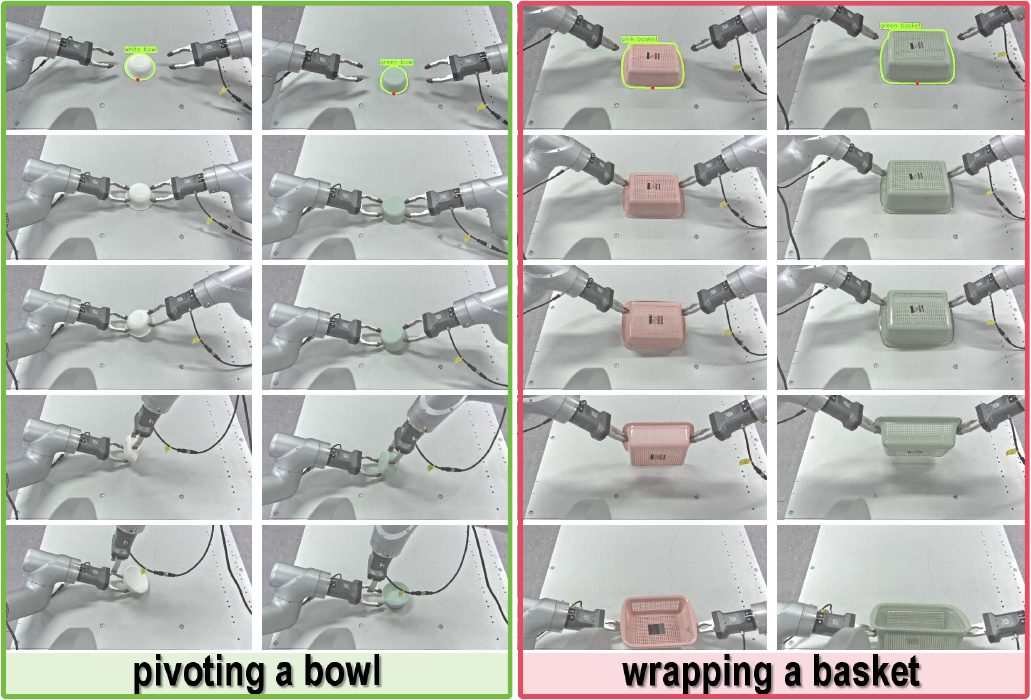}
	\vspace{-18pt}
	\caption{Qualitative real robot rollout samples of two bimanual non-prehensile skills (\textit{pivoting} and \textit{wrapping}) in another novel dual-arm manipulator platform.}
	\label{crossEmbodiment}
	\vspace{-15pt}
	\end{center}
\end{figure}

\begin{figure}[t]
	\begin{center}
	\includegraphics[width=1.0\linewidth]{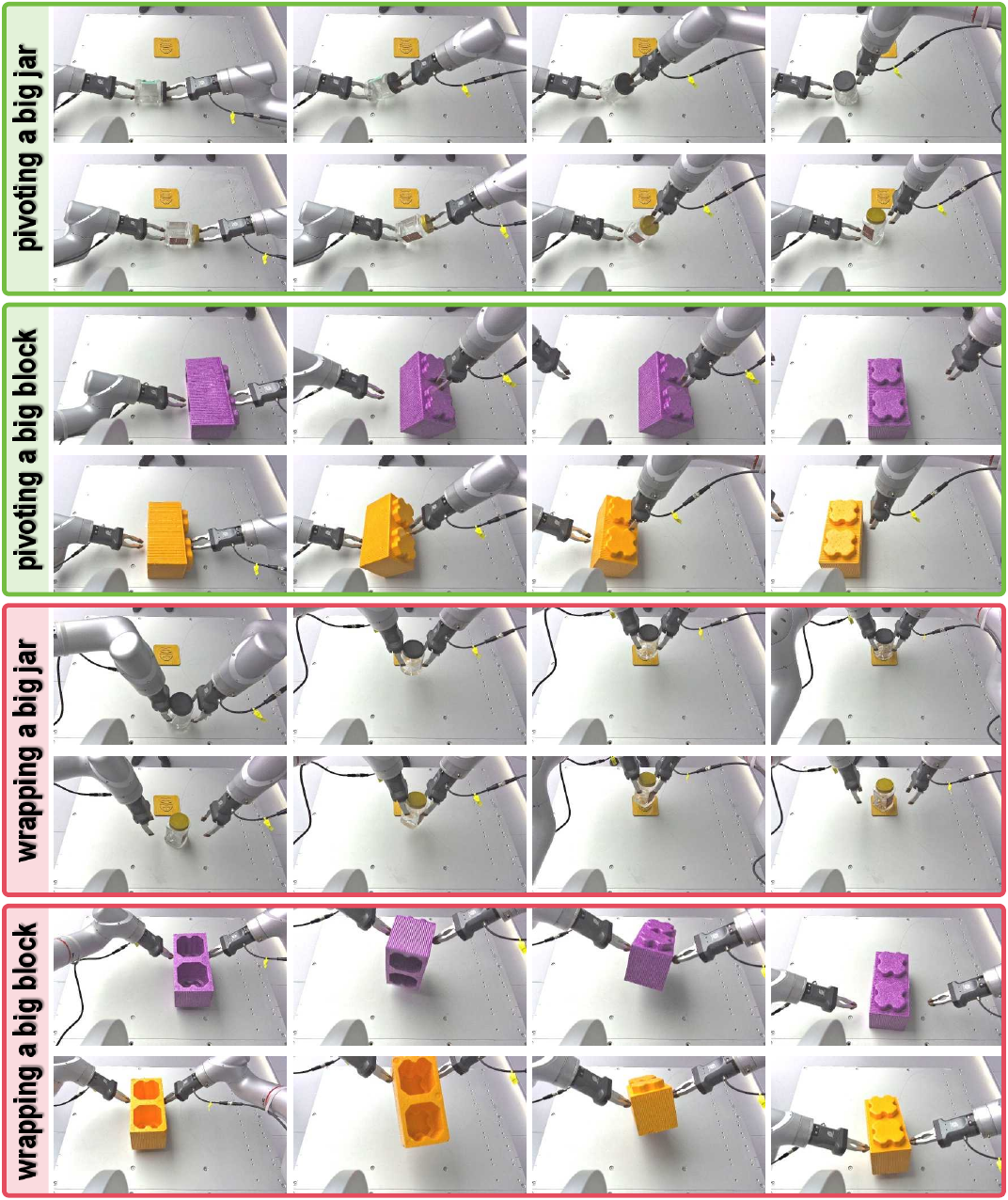}
	\vspace{-18pt}
	\caption{Visualization results of new skill transfer experiments performed on the humanoid dual-arm robot. We demonstrated that BiNoMaP can be applied to new objects (e.g., the big jars and big blocks) with shapes completely different from assets we used before. Best to view after zooming in.}
	\label{newAdaptability}
	\vspace{-15pt}
	\end{center}
\end{figure}

\begin{figure*}[t]
	\begin{center}
	\includegraphics[width=1.0\linewidth]{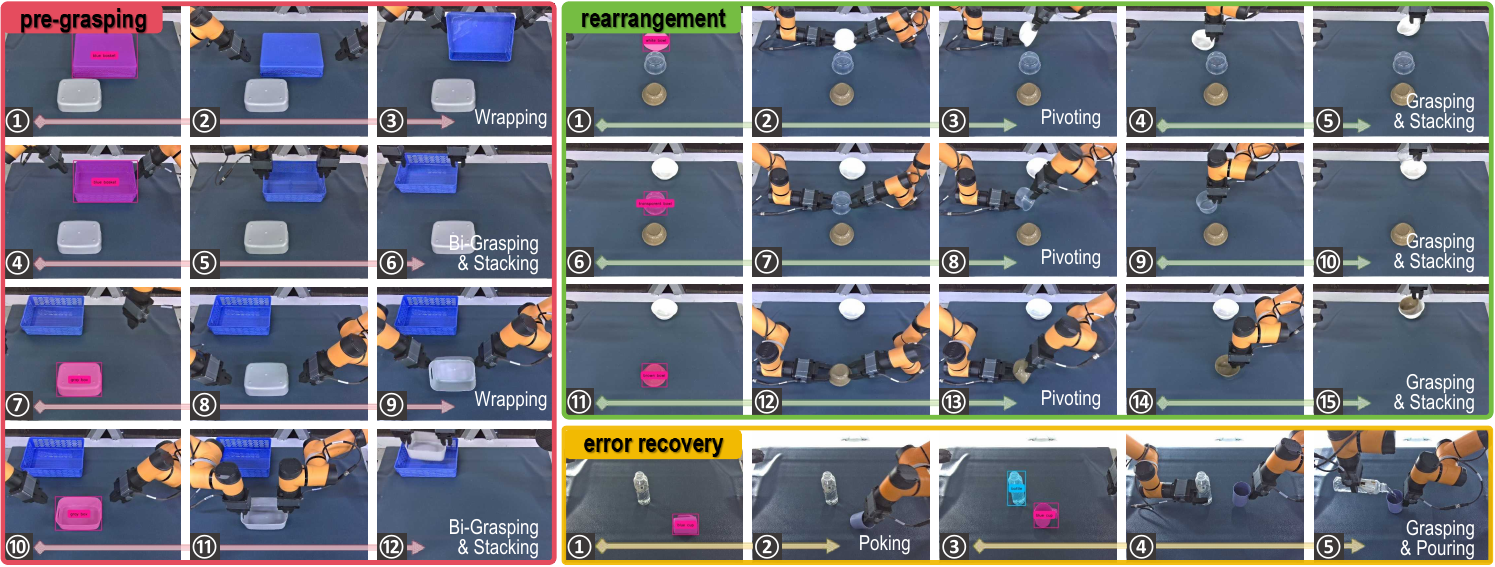}
	\vspace{-18pt}
	\caption{Examples of utilizing both ready-made grasping skills and learned non-prehensile skills to boost complex long-horizon manipulation tasks. These three downstream examples serve as preliminary demonstrations of the application potential of bimanual non-prehensile manipulation.}
	\label{applications}
	\vspace{-15pt}
	\end{center}
\end{figure*}

\subsubsection{Embodiment-Agnostic Skill Representation \textsf{(A4)}} 
To further demonstrate the usability and convenient transferability of BiNoMaP, we evaluate whether the learned manipulation primitives can be directly migrated across different robotic embodiments without re-training. Specifically, we validate BiNoMaP on another unseen dual-arm robotic platform configured in a humanoid style (hardware details are shown in Fig.~\ref{platforms}). For this evaluation, we transfer two learned skills (\textit{pivoting a bowl} and \textit{wrapping a basket}), and select several corresponding object assets for each task. Aside from the necessary adjustment of the XYZ axis order (due to the humanoid-style mounting differing from the primary opposing-arm setup), no significant engineering effort is required. Importantly, unlike prevailing visuomotor diffusion policies \cite{zhao2023learning, chi2023diffusion, ze2024dp3} or RL-based methods \cite{zhou2023hacman, cho2024corn, lyu2025dywa}, BiNoMaP does not necessitate recollecting demonstrations or building custom simulation assets for retraining. As shown in Fig.~\ref{crossEmbodiment}, the system successfully performs human-like actions of flipping bowls and baskets upright, highlighting the cross-embodiment capability of BiNoMaP. These results further underscore its promising potential for real-world deployment in diverse robotic hardware configurations.

 As previously demonstrated, BiNoMaP is not only effective for objects with substantial intra-class geometric variation but can also transfer skills across novel object categories with drastically different shapes. For the \textit{pivoting} skill, our original experiments on the primary paltform already covered three geometrically distinct objects: a bowl (hemisphere), box (cuboid), and bottle (cylinder). For the \textit{wrapping} skill, we demonstrated transferring across a ball (sphere) and basket (cuboid). To further highlight BiNoMaP’s practical usability and strong cross-object geometric generalization, we conducted additional experiments on the humanoid dual-arm platform. Specifically, the \textit{pivoting} skill was transferred to two new kinds of objects with significant shape differences (big jars and big blocks). The \textit{wrapping} skill was similarly transferred to big jars and big blocks, expanding beyond the ball and basket used in the initial evaluation. Representative visualizations of these new trials are provided in in Fig.~\ref{newAdaptability}. We believe that these results further validate BiNoMaP’s generalization when there have significant variations in shape, size, or mass distribution of new unseen objects.

\subsubsection{Compositionality for Downstream Tasks \textsf{(A5)}}
Another key advantage of BiNoMaP lies in the modularity and transferability of its learned atomic skills, which can be seamlessly integrated into higher-level downstream manipulation tasks. We explore three representative applications that highlight this property: (1) using the \textit{wrapping basket} skill to flip an inverted basket upright for subsequent bimanual grasping (\textbf{pre-grasping}); (2) applying the \textit{pivoting bowl} skill to sequentially flip and stack multiple inverted non-prehensile bowls on the table (\textbf{rearrangement}); and (3) employing the \textit{poking mug} skill to upright a fallen mug so that filling water poured from a bottle can be performed afterward (\textbf{error recovery}). In each case, once the object is manipulated into a graspable state by BiNoMaP, we leverage VLMs to localize and segment the object, apply AnyGrasp \cite{fang2023anygrasp} to obtain 6-DoF grasp poses, and execute motion planning via inverse kinematics. As illustrated in Fig.~\ref{applications}, all three cases were validated in real-world experiments under the same hardware setup, yielding consistently successful results. These demonstrations underscore the composability and practical utility of BiNoMaP, and point toward promising future directions where the framework can be embedded into longer-horizon and more complex interactive robotic manipulation pipelines.

\begin{figure*}[t]
	\begin{center}
	\includegraphics[width=1.0\linewidth]{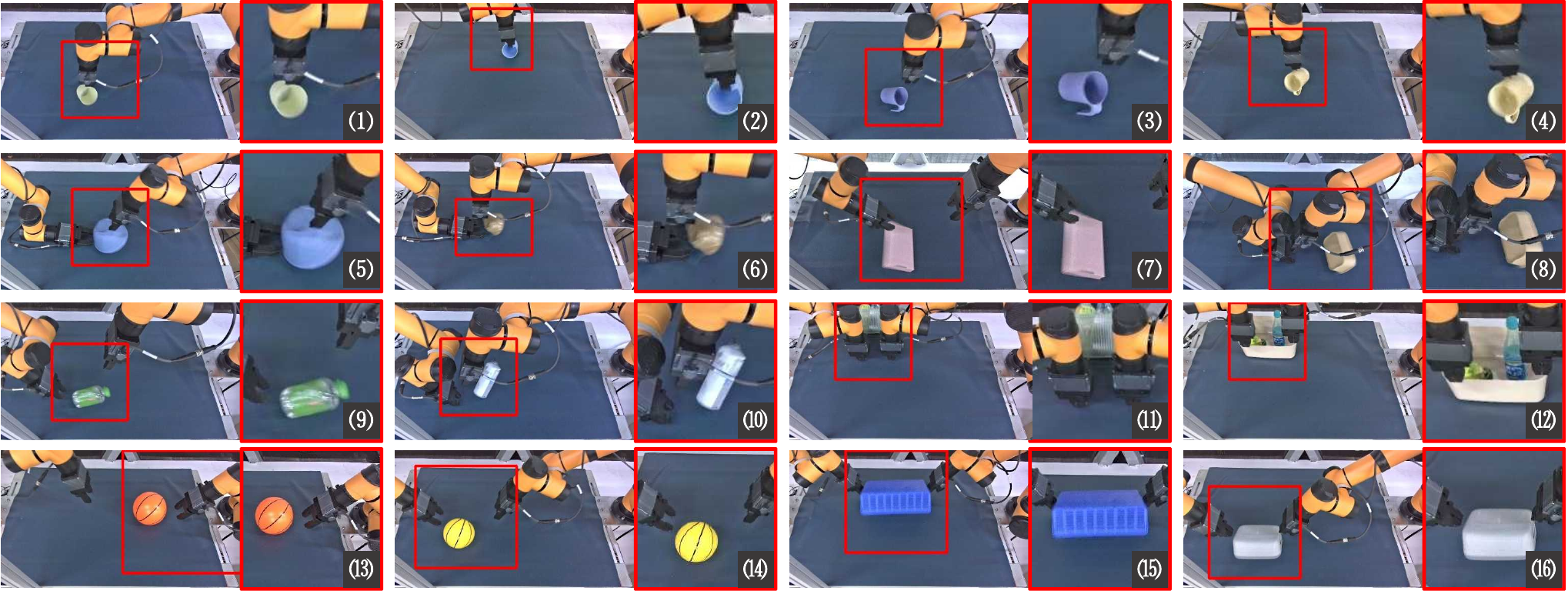}
	\vspace{-18pt}
	\caption{Examples of failed cases in all four skills (\textit{poking}-(1,2,3,4), \textit{pivoting}-(5,6,7,8,9,10), \textit{pushing}-(11,12), and \textit{wrapping}-(13,14,15,16)) and eight tasks during real robot evaluation. We have outlined and magnified the areas where the failures occurred so that we can quickly examine them.}
	\label{failures}
	\vspace{-15pt}
	\end{center}
\end{figure*}

\subsection{Summary and Analysis of Failure Cases}
To better understand the practical performance of BiNoMaP, we collected and summarized representative failure cases observed in real-robot experiments across the four skills and eight tasks, with visualizations provided in Fig.~\ref{failures}. We explain the main factors that cause these errors as follow. 

(\textbf{\romannumeral 1}) For the \textit{poking} skill, failures occurred when lifting uniformly distributed handless cups due to insufficient insertion of the gripper into the cup mouth, leading to unstable placements and eventual toppling (Fig.~\ref{failures} (1,2)), and when handling mugs with uneven mass distribution, where the handle frequently caused rotational displacements under gravity, significantly increasing failure rates due to dynamic and unpredictable changes (Fig.~\ref{failures} (3,4)).

(\textbf{\romannumeral 2})  For the \textit{pivoting} skill, failures were often observed when flipping bowls, where either insufficient (Fig.~\ref{failures} (5)) or excessive (Fig.~\ref{failures} (6)) contact between the right arm's end-effector and the bowl caused slipping or bouncing. When raising thin rectangular boxes, the box either over-tilted toward the left arm (Fig.~\ref{failures} (7)) and continued falling after release—suggesting the need for corrective post-actions—or was destabilized by excessive end-effector contact (Fig.~\ref{failures} (8)). And when lifting bottles, similar issues of under-contact (Fig.~\ref{failures} (9)) or over-contact (Fig.~\ref{failures} (10)) with the bottle's head and bottom resulted in slipping or bouncing failures.

(\textbf{\romannumeral 3}) For the \textit{pushing} skill, asynchronous motion between the two arms during basket pushing led to tipping of the heavy objects inside (Fig.~\ref{failures} (11,12)).

(\textbf{\romannumeral 4}) For the \textit{wrapping} skill, failures in ball manipulation included arm self-collision (Fig.~\ref{failures} (13)) and asynchronous release leading to significant moving during placement (Fig.~\ref{failures} (14)), where the latter is considered an unstable outcome. In the basket-wrapping task, the most frequent failure mode was slippage during lifting and flipping (Fig.~\ref{failures} (15,16)), primarily caused by minor discrepancies in the inter-arm distance, which is inherently difficult to maintain perfectly.

\begin{figure}[t]
	\begin{center}
	\includegraphics[width=\linewidth]{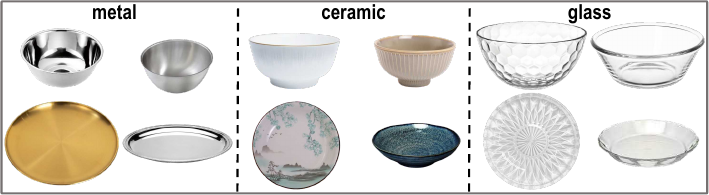}
	\vspace{-18pt}
	\caption{Difficult object examples that our BioMaP cannot solve at present, especially when these objects are upside down on the table and need to be flipped via applying the bimanual \textit{pivoting} skill.}
	\label{hardCases}
	\vspace{-15pt}
	\end{center}
\end{figure}

\textbf{More Challenging Non-Prehensile Cases}: The current BiNoMaP framework, which does not incorporate force–torque sensing, is unable to manipulate objects with extremely low tolerance to errors, particularly in \textit{pivoting} related tasks. Typical examples include smooth ceramic or metal bowls, ultra-thin ceramic or metal plates, and fragile glassware, as illustrated in Fig.~\ref{hardCases}. These rigid objects cannot withstand even visually perceptible deformations, making it challenging to balance contact distance and contact force for successful flipping, even if force sensing were available. We also tested BiNoMaP on these challenging objects, but all of them failed. A potential direction to address this limitation is the use of multi-fingered dexterous hands for more delicate manipulation \cite{li2025dexnoma}, or alternatively leveraging the external dexterity such as table edges \cite{wang2025dexterous} to facilitate stable interactions.

\section{Conclusion}\label{sec:conclusion}

In this work, we introduced BiNoMaP, an RL-free framework that formulates bimanual non-prehensile manipulation as geometry-structured skill learning. By transforming coarse human demonstrations into contact-aware, parameterized motion primitives, BiNoMaP enables category-level generalization across diverse object instances. Extensive real-world experiments demonstrate consistent improvements over strong visuomotor and RL-based baselines across multiple skills and objects. Importantly, because the primitives are defined in task space via geometry-constrained relative poses rather than embodiment-specific policies, they transfer naturally across dual-arm robotic platforms with different kinematic structures. Together with their composability for downstream tasks, these results highlight BiNoMaP as a scalable and transferable approach to non-prehensile bimanual manipulation.

\textbf{Limitations:} First, the current overall framework operates in an open-loop manner and lacks closed-loop error correction, making it sensitive to execution deviations. Second, without tactile sensing, it is less effective for strictly rigid objects where precise force feedback is critical to achieve the dexterous manipulation. Finally, BiNoMaP remains skill- or category-specific, integrating it into a unified language-driven VLA system for end-to-end multi-skill scheduling is an important direction for future work.

{ \small  
\bibliographystyle{plainnat}
\bibliography{refs}
}

\end{document}